\documentclass[10pt, conference, compsocconf]{IEEEtran}

\usepackage{graphicx}
\usepackage[section]{placeins}
\usepackage{float}
\usepackage[export]{adjustbox}
\usepackage{mathtools}
\usepackage[T1]{fontenc}
\usepackage[utf8]{inputenc}     
\usepackage[english]{babel}      
\usepackage{hyperref}

\usepackage{amsmath}

\usepackage{algorithmic}
\usepackage{array}

\ifCLASSOPTIONcompsoc
  \usepackage[caption=false,font=footnotesize,labelfont=sf,textfont=sf]{subfig}
\else
  \usepackage[caption=false,font=footnotesize]{subfig}
\fi

\usepackage{url}

\usepackage{microtype}

\author{\IEEEauthorblockN{Ramkumar B, R. S. Hegde}
\IEEEauthorblockA{Department of Electrical Engineering\\
Indian Institute of Technology\\
Gandhinagar, Gujarat, India 382355\\
Email: hegder@iitgn.ac.in}
\and
\IEEEauthorblockN{Rob Laber, Hristo Bojinov}
\IEEEauthorblockA{Innit Inc. \\
Redwood City, CA 94063, USA
}}

\begin{document}
%
\title{GPGPU Acceleration of the KAZE Image Feature Extraction Algorithm}



%


\maketitle

\begin{abstract}

 The
recently proposed open-source KAZE image feature detection and description
algorithm~\cite{Alcantarilla2012} offers unprecedented performance in comparison to conventional ones like SIFT and SURF as it relies on nonlinear scale spaces instead of
Gaussian linear scale spaces. The improved performance, however, comes with a
significant computational cost limiting its use for many applications. We report a GPGPU implementation of the KAZE algorithm without resorting to binary descriptors for gaining speedup. For a 1920 by 1200 sized image our Compute Unified Device Architecture (CUDA) C based GPU version took around 300 milliseconds on a NVIDIA GeForce GTX Titan X (Maxwell Architecture-GM200) card in comparison to nearly 2400 milliseconds for a multithreaded CPU version (16 threaded Intel(R) Xeon(R) CPU E5-2650 processsor). The CUDA based parallel implementation is described in detail with fine-grained comparison between the GPU and CPU implementations. By achieving nearly 8 fold speedup without performance degradation our work expands the applicability of the KAZE algorithm. Additionally, the strategies described here can prove useful for the GPU implementation of other nonlinear scale space based methods.

\end{abstract}



%
\IEEEpeerreviewmaketitle

\section{Introduction} \label{sec:intro} 

Feature point detection~\cite{Lindeberg1998} and description~\cite{Bianco2015} is a key tool for many computer vision based applications, such as visual navigation~\cite{Gauglitz2011}, automatic target recognition~\cite{Sanna2014}, tracking, structure from motion, registration, calibration, and more. By picking out only the most salient points of an image, that can be repeatably localized across different images, we can vastly reduce subsequent data processing. Feature extraction~\cite{Lindeberg1998}, however, still remains a major bottleneck for many implementations due to the high computational cost, especially those that are most robust. GPGPUs have become promising for image processing~\cite{Fung2008}, computer vision~\cite{Seung2008,Fung2008}, and deep learning tasks~\cite{Coates2013}.  The GPU is an appropriate choice to process image related problems because these involve large
data sizes and exhibit a high arithmetic intensity (the ratio between arithmetic operations and memory operations). 

Scale Invariant Feature Transform (SIFT)~\cite{Lowe2004,Mikolajczyk2005,Bianco2015} is widely considered as one of the most robust feature descriptors that exhibits distinctiveness and invariance to common image transformations. 
 Some of the binary descriptors  have been developed to detect and match interest regions that can decrease the computational cost. FAST~\cite{Rosten2006} key-point detectors with Binary Robust Independent Elementary Features (BRIEF) feature descriptor~\cite{Calonder2010} have yielded better results in real time applications. However, BRIEF and other such binary descriptors are not very robust to image transformations. Binary Robust Invariant Scalable Keypoints (BRISK)~\cite{Leutenegger2011} and Oriented FAST and rotated BRIEF (ORB)~\cite{Rublee2011}  methods have been developed by modifying  BRIEF and FAST to achieve scale and rotation invariance to some degree. The recently proposed KAZE~\cite{Alcantarilla2012} algorithm outperforms SIFT and several other algorithms inspired by it. In the short span of its introduction, it has already become widely used in image matching~\cite{Lehiani2016}, target classification~\cite{Gao2017}, and data mining~\cite{Camargo2014,Zhai2014}. 

\begin{figure*}[htbp]
\centering
\includegraphics[width=0.6\textwidth]{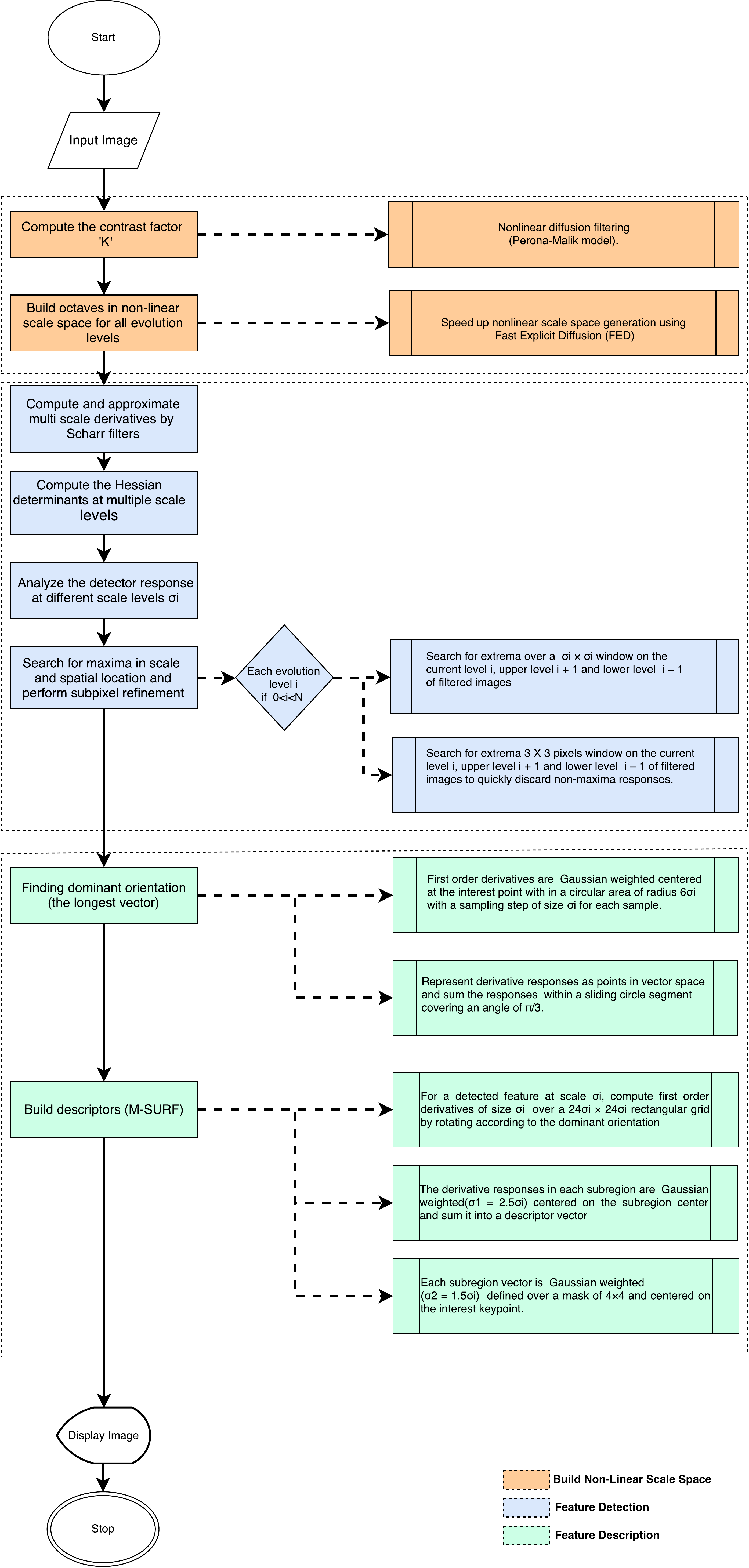}
\caption{Flowchart for KAZE Algorithm}
\label{fig:flowchart}
\end{figure*}

The improved performance of the KAZE algorithm~\cite{Bianco2015} however comes at the cost of increased computational cost.  The original KAZE algorithm provides modified SURF (Speeded up robust features) like descriptors allowing it to be used as a replacement for SIFT and SURF. An accelerated version (multithreaded CPU version) of the original KAZE algorithm termed as A-KAZE~\cite{Alcantarilla2013} was proposed by the original authors and reduces the computational cost at the expense of algorithm performance in some scenarios. The A-KAZE uses binary descriptors in place of SURF-like descriptors and thus may not easily replace SURF in computer vision pipelines. Subsequently, the authors have reported on a GPU based implementation of the A-KAZE~\cite{Pieropan2016}. Although this has yielded better performance in terms of computational cost when compared to CPU based KAZE, a compromise of algorithm performance in certain outcomes is again involved due to it resorting to binary descriptors. 

In this paper, we mainly focus on parallelizing the original version of KAZE on the GPU (using Compute Unified Device Architecture (CUDA) C) without resorting to binary descriptors. We primarily take advantage of the characteristic of the algorithm and find out parallel strategies for each stage of KAZE algorithm. Our implementation is primarily GPU based; the CPU just plays a part in controlling the GPU, initializing CUDA kernels and data allocation and is free to perform other tasks during the bulk of the computation.  The paper is organized as follows, following this introduction the KAZE algorithm and its parts are
described in detail in~\autoref{sec:kaze}. In~\autoref{sec:implementation}, we have discussed the implementation of KAZE algorithm on GPU using CUDA. The obtained results have been discussed in the~\autoref{sec:results}. Section~\ref{sec:conclusion} concludes the paper.

\section{The KAZE algorithm}
\label{sec:kaze}

The KAZE Features~\cite{Bianco2015} algorithm is a novel feature detection and
description method and it belongs to the class of methods which utilize the so-called scale space. Its novelty arises in that it operates using
a nonlinear scale space whereas previous methods such as SIFT or SURF
~\cite{Bay2008,Bianco2015} find features in the Gaussian scale space (a particular
instance of linear diffusion). The three main steps involved in KAZE feature extraction
algorithm are: 

\begin{enumerate}
  \item Construct a Nonlinear Scale space pyramid of the original
  image. 
\item Determine keypoints using Hessian determinants and multiscale
    derivatives in the nonlinear scale space.   
  \item Compute orientation and descriptor vectors for all keypoints. 
  \end{enumerate} 
  
The detailed overview of these steps in the KAZE algorithm is given in~\autoref{fig:flowchart} and further explained below.

\subsection{Building the nonlinear scale space}
The main idea for constructing scale space is to obtain a separable structures of image 
from the original image, such that only fine scale
image structures exist in the multi-scale representation~\cite{Lindeberg1998,Harvey1996} such as scale-space
representation, pyramids, and non-linear diffusion methods. Linear scale space systems~\cite{Harvey1996} (like the Gaussian) smear
the edges such that small-scale segmentations do not coincide with large-scale segmentations leading to reduction in localization accuracy. Non linear
scale space systems overcome this by locally blurring the image data so that
details remain unaffected and noise gets blurred. But computing the nonlinear scale spaces are slow and
diffusion function should be chosen carefully . 

Consider an image of size $W \times H$ (width and height of image respectively). Anisotropic diffusion filter has been utilized while smoothing in order to preserve the edges and is defined as: 
\begin{equation}
  \label{eq:anis}
\frac{\mathrm d I}{\mathrm d t} = div\left( c(x,y,t). \nabla I \right),
\end{equation}
where $I$ is the image luminance, div is the divergence operator,
$c(x,y,t)$ is the
conductivity function which plays a key role in controlling the diffusion of the
image, $\nabla$ is the gradient operator and $t$ is the timing scale
parameter. Magnitude of the image gradient controls the diffusion. The
conductivity function is defined as follows:
\begin{equation}\label{eq:conduc}
c(x,y,t) = g(|\nabla I_{\sigma}((x,y,t)|),
\end{equation}
where the function $ \nabla I $ is the image gradient obtained after applying
Gaussian smoothing on the  image.~\autoref{eq:anis} is generally known
as Perona and Malik diffusion equation~\cite{Perona1990}. The two different forms
of conductivity function '$g$' described by Perona and Malik are:
\begin{equation}\label{eq:perona}
g1=exp\left(-  \frac{\mathrm |\nabla I_{\sigma}|^2}{\mathrm k^2} \right), g2= \frac{\mathrm 1}{\mathrm 1 + \frac{\mathrm |\nabla I_{\sigma}|^2}{\mathrm k^2}},
\end{equation}
where '$k$' is the contrast factor that decides whether the edges to be smoothed
or filtered out.

We need to approximate the differential equations using
numerical methods as there are no analytical solutions for~\autoref{eq:anis}. The simple solution is to use explicit schemes but these are found
to be impractical for feature detection due to their computational complexity. To accelerate the nonlinear scale space generation, Additive Operator Fast
Explicit Diffusion (FED)~\cite{Feng2016,Grewenig2010} scheme was introduced by
~\cite{Alcantarilla2012}. The nonlinear scale space can be built efficiently by means of such
Fast Explicit Diffusion (FED)~\cite{Grewenig2010} schemes and they are numerically stable for
any step size.

The schemes discretize the~\autoref{eq:anis}
as follows:
\begin{equation}\label{eq:discanis}
\frac{I^{i+1} - I^i}{\tau} = \displaystyle\sum_{l=1}^{m} A_{l}(I^i)I^{i+1},
\end{equation}
where $ A_{l} $ encodes the image conductivities for each
dimension, $ \tau $ is a constant time step to maintain stability.
The main intention of FED schemes is to perform $M$ cycles of $n$ explicit diffusion
steps with varying step sizes $ \tau_{j} $ and is defined by:
\begin{equation}\label{eq:fed}
\tau_{j}=\frac{\tau_{max}}{2\cos^2\left(\pi \frac{2j+1}{4n+2}\right)}.
\end{equation}

Discretize the scale space in logarithmic steps that are arranged in a series of $O$
octaves and $S$ sub-levels. We always preserve the resolution of original image at
each new octave whereas in SIFT we perform downsampling at each new octave. 
Discrete values of octave index $o$ and a sub-level index $s$ are used to identify
set of octaves and sub-levels respectively. The octave and the sub-level indexes
are mapped to their corresponding scale $\sigma$ as defined by the following
equation:
\begin{eqnarray}
  \label{eq:sigma}
\displaystyle \sigma_{i}(o,s) = \sigma_{0}2^{\frac{o+s}{S}}, \\ \nonumber 
o \in [0...O-1], s \in [0...S-1], \\ \nonumber
i\in[0...N],
\end{eqnarray}
where $\sigma_{0}$ is the base scale level and $N$ is the total number of filtered
images. 

Pixel units $ \sigma_{i}$ in the  set of discrete scale levels is
converted to time units because of nonlinear diffusion filtering which is
defined in units of time. In the Gaussian scale space, filtering the image for
some time $t$ = $\frac{\sigma^2}{2}$ is equivalent to convolution of the image
with Gaussian of standard deviation $\sigma$ (in pixels). The result of this
conversion is applied to transform the scale space $\sigma_{i}(o, s)$ in units
of time and obtain a set of evolution times by mapping $\sigma_{i} \rightarrow t_{i} $
defined by:
\begin{equation}\label{eq:evolve}
t_{i}=\frac{1}{2} \sigma_{i}^2, i={0...N}.
\end{equation}
Nonlinear scale space is constructed from the obtained set of evolution times
using $\sigma_{i} \rightarrow t_{i} $ mapping. For each filtered image $t_{i}$ in the non
linear scale space, the convolution of the original image with a Gaussian of
standard deviation $\sigma_{i}$ does not correspond with the resulting image.

\subsection{Feature Detection}

Feature detection~\cite{Li2015,Gauglitz2011} is the identification
of interesting image primitives (e.g. points, lines/curves, and regions) for the
purpose of highlighting salient visual cues in digital images. The primary goal of feature detection is to effectively extract highly stable visual features.

In order to increase the detection accuracy, Hessian determinant is computed for each filtered image $L_{i}$ in the nonlinear scale space. $k_{i,norm}
=\frac{k_{i}}{2^{o^i}} $ is a scaling factor to normalize the computed Hessian
determinant for the corresponding filtered images. 
 \begin{equation}\label{eq:hessian}
 L_{i,Hessian} = k_{i,norm}^2(L_{i,xx}L_{i,yy} - L_{i,xy}L_{i,xy}).
 \end{equation}
 
Concatenated Scharr filter of step size $k_{i,norm}$ is used to compute second
order derivatives. When compared to other filters, Scharr
filters~\cite{Weickert2002} have  better central differences differentiation and
rotation invariance. First, we search for detector response maxima in spatial
location. Preserve the maxima in $3 \times 3$ pixels window with detector responses
higher than the threshold for each step. We cross-check whether the response is
maxima with respect to other keypoints from level $i-1$ to $i+1$ for each of the
response directly below and above  in a $\sigma_{i,s}\times \sigma_{i,s} $ pixels
window. Hessian determinant response in $3 \times 3$ pixels neighborhood is fitted by
the two-dimensional quadratic function and its maximum value is used to estimate the two-dimensional
position of the keypoint with sub-pixel accuracy.
\subsection{Feature Description} A feature
description is a process which takes an image with interest points and yields
feature descriptors (vectors). The obtained feature descriptors acts as numerical "fingerprint" which differentiates one feature from another and are ideally invariant to image transformations. There are several methods to find feature
descriptors like SIFT, SURF, Histogram Oriented Gradients
(HOG)~\cite{Dalal2005}. Here we use modified SURF (M-SURF)~\cite{Huang2010} for
computing descriptors. 
\subsubsection{Finding the Dominant Orientation}
Estimating the dominant orientation with keypoint location as a center in a local
neighborhood is a key step for obtaining rotation invariant descriptors. We
compute the dominant orientation with a sampling step of size $\sigma_{i}$ in a circular area of radius $6\sigma_{i}$. First order derivatives $L_{x}$ and
$L_{y}$ are weighted with a Gaussian weighted at the keypoint location as
a center for each of the samples in the circular area. Then, represent the
derivative responses as points in vector space and sum the responses within
a sliding circle segment covering an angle of $\pi/3$ to find the dominant
orientation. Finally, dominant orientation is obtained from the longest vector.

\subsubsection{Building the Descriptor} Frame work of nonlinear scale space is
embedded with M-SURF for computing descriptors. Compute the first order
derivatives $L_{x}$ and $L_{y}$ of size $\sigma_{i}$  over a $24\sigma_{i} \times 24\sigma_{i}$ rectangular window for a detected feature at scale $\sigma_{i}$. This rectangular window is divided into $4 \times 4$ subregions with an overlap of
$2\sigma_{i}$ and with the size $9\sigma_{i} \times 9\sigma_{i}$. Gaussian
weighted ($\sigma_{1} = 2.5\sigma_{i}$) derivative responses in each subregion is
centered on the subregion center and summed into a descriptor vector $d_{v} = (
L_{x} , L_{y} , |L_{x} |, |L_{y} |)$. Then, each subregion vector is Gaussian
weighted ($\sigma_{2} = 1.5\sigma_{i}$ ) over a mask of $4 \times 4$ with keypoint as
center. Compute the the derivatives according to the dominant orientation. In order to achieve the contrast invariance, normalize the
descriptor vector of length 64 into a unit vector.

\section{CUDA Implementation of the KAZE algorithm}
\label{sec:implementation}

While the KAZE algorithm has demonstrated superior performance in terms of
feature robustness in relation to the class of linear scale space based methods,
the computations involving the nonlinear scale space considerably add to the
computational cost. However, a lot of parallelism exists in the algorithm and it
is thus suitable for a GPU based acceleration. We describe the CUDA based implementation
details for all the three main steps of the KAZE algorithm in this section. As long as they are available we always use standard OpenCV GPU primitives. 

We have used Gpumat object of OpenCV to allocate memory in GPU global memory. We allocate memory to evolution level with dimensions $O \times S$ (number of octaves and number of sublevels respectively) for each evolution image of nonlinear scale space pyramid ($L_{t}$), first order derivatives ($L_{x}$ and $L_{y}$), second order derivatives ($L_{xx}$, $L_{yy}$ and $L_{xy}$), images obtained after Gaussian smoothing ( $L_{smooth}$) and hessian determinants ($L_{det}$) of size $W \times H$. 

\subsubsection{Nonlinear scale space generation}
For a given input image, in order to reduce image artifacts and noise, we convolve the image with a Gaussian kernel of standard deviation $\sigma_{0}$. The resultant image is then considered as the base image and its image gradient histogram is
computed to obtain the contrast parameter $k$ in an automatic procedure. 
For the obtained set of evolution times $t_{i}$ and contrast parameter, we then build the
nonlinear scale space using the FED scheme.
At each evolution level of the nonlinear scale space, several auxiliary images are generated. The image 
$L_{smooth}$ is the image obtained by performing a two-dimensional Gaussian convolution using OpenCV CUDA primitive kernel on the image generated at the end of the previous evolution level (or on the base image in the case of the first level). $L_{smooth}$ is calculated by allocating a block of $8 \times 8$ threads and a grid of $(W+8)/8 \times (H+8)/8$ blocks. First order $x$ and $y$ Gaussian derivatives of $L_{smooth}$ in the form of images  $L_{x}$ and $L_{y}$ are then calculated using Scharr filters~\cite{Weickert2002}. We determine the diffused image $L_{flow}$ at each level by then using our GPU kernel to compute Perona malik conductivity equation from the first order Gaussian derivatives $L_{x}$ and $L_{y}$. We allocate a block of $32 \times 32$ threads and a grid of $(W+32)/32 \times (H+32)/32$ blocks and then each thread computes the conductivity equation from the first order Gaussian derivatives on each pixel of an image to form the flow image. From the obtained flow images, we need to determine step images. So, we implemented Fast Explicit Diffusion that executes over several steps (for defined step size) using a GPU kernel that yields step image at each evolution level. We have allocated a block containing $32 \times 32$ threads and a grid of $(W+32)/32 \times (H+32)/32$ blocks so that each thread computes Fast Explicit Diffusion Scheme (FED) using diffusivity factor of image over the inner steps at each scale to build nonlinear scale space. The obtained step image after applying FED scheme is considered as the image of the corresponding evolution level. The performance of nonlinear scale space generation kernel has been improved using shared memory access when appropriate. 

\subsubsection{Implementation of Feature Points Detection}
In order to determine the detector response at each evolution level, we calculate the multiscale derivatives. To find the first order and second order derivatives at each evolution level for larger scale size, we use separable larger linear row filter and the larger linear column filter kernels and combine them for calculating Scharr derivatives. We have allocated a block containing $64 \times 4$ threads and a grid of $(W+64)/64 \times (H+4)/4$ blocks for both row filter and column filters so that each thread computes the derivative response of each pixel of the image at each evolution level of defined kernel window with scale size $\sigma_{i}$. We combine the results of both the filter to form large and separable linear two dimensional filter which yields multiscale derivative responses of the image at each evolution level with scale size $\sigma_{i}$. From the obtained second order derivatives $L_{xx}$,$L_{yy}$ and $L_{xy}$, we calculate detector response of images at each evolution level. Although we compute multiscale derivatives for every pixel in the feature detection step, we also utilize the same set of computed derivatives in the feature description step in order to reduce computational cost. We allocate $8 \times 8$ threads in a two-dimensional block. First, the current and two adjacent scales in non-linear space should be bound to three texture references. Each thread takes out 27 corresponding pixels in three levels by global fetch. Then, for each of the potential response, we check that the response is a maxima with respect to other keypoints from level $i - 1$ and $i + 1$, respectively directly above and directly below in a window of size $\sigma_{i,s} \times \sigma_{i,s} $ pixels. The central point in these pixels is compared to others to determine whether it is the extreme point. If a pixel is an extreme point, it is selected as a candidate keypoint. Last, the edge responses must be eliminated. To eliminate edge points, the principal curvature around the surface $D(x,y)$ of a candidate point at (x,y) can be calculated using the Hessian matrix of candidate keypoints. In practice, the image pixels are discrete, so the Hessian matrix can be calculated by 8 points around those points. The Hessian matrix is given by
\begin{equation}\label{eq:hessmat}
H = 
 \begin{pmatrix}
  D_{xx} & D_{xy}  \\
  D_{xy} & D_{yy}  \\
 \end{pmatrix}
\end{equation}
Where $D_{xx}$, $D_{xy}$, and $D_{yy}$ are second order local derivative of candidate keypoint. Let the largest eigenvalue be $\alpha$, and the smallest eigenvalue is $\beta$. Then the sum of the eigenvalues from the trace of H and their product from the determinant can be computed as:
\begin{equation}\label{eq:10}
Tr(H)=D_{xx}+D_{xy}=\alpha + \beta 
\end{equation}
\begin{equation}\label{eq:hessdet}
Det(H)=D_{xx}D_{yy}-D_{xy}^2
\end{equation}
Let r be the ratio between the $\alpha$ and $\beta$, so $\alpha=r \beta$. Then,
\begin{equation}\label{eq:hessfin}
\frac{Tr(H)^2}{Det(H)} = \frac{(\alpha + \beta)^2}{\alpha\beta} = \frac{(r+1)^2}{r},
\end{equation}

\begin{figure*}[htbp]
\centering
\includegraphics[width=\textwidth]{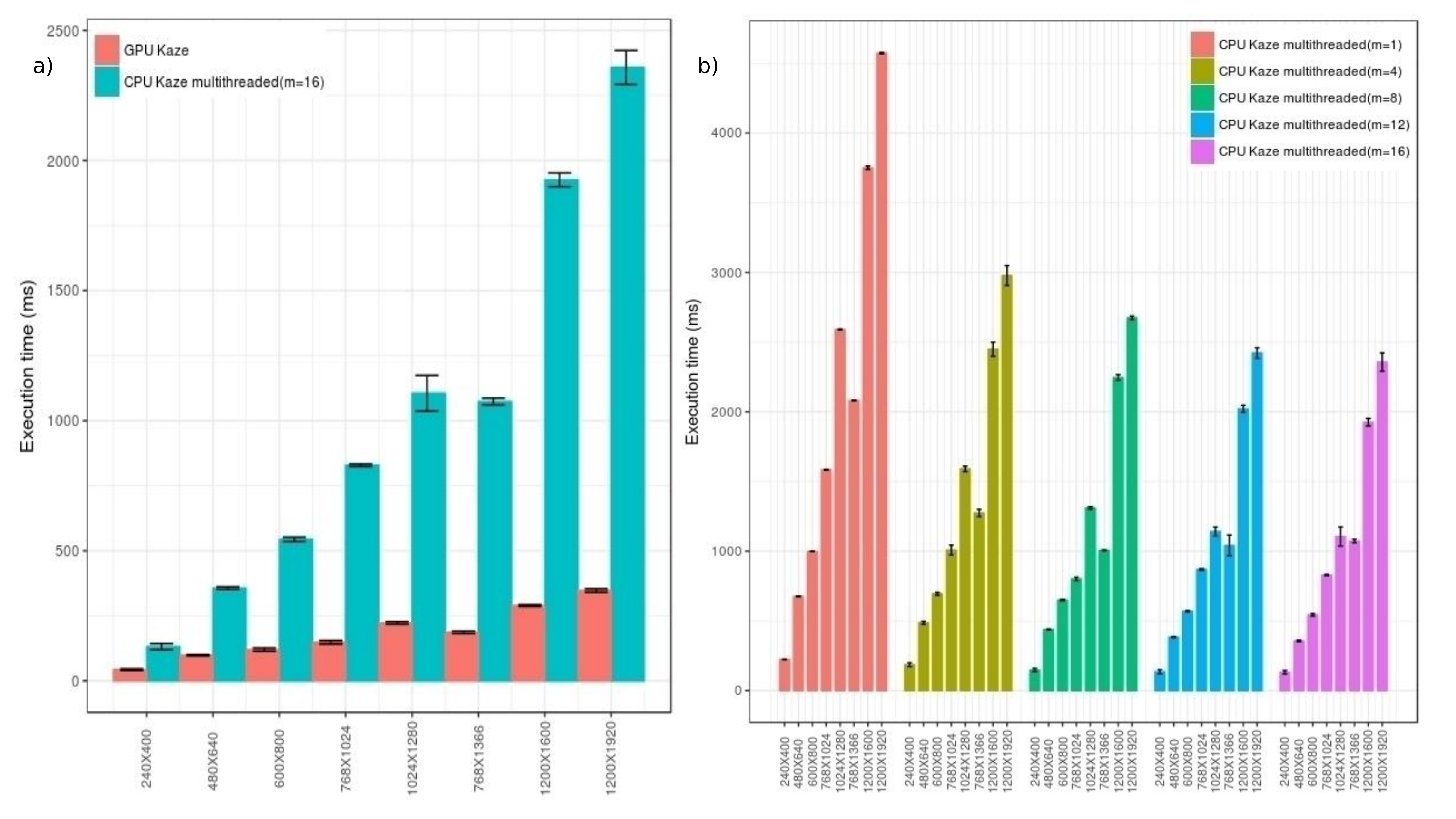}
\caption{Runtimes of multithreaded CPU-KAZE and GPU-KAZE}
\label{fig:Timing performance}
\end{figure*}

\begin{figure*}[htbp]
\centering
\includegraphics[width=\textwidth]{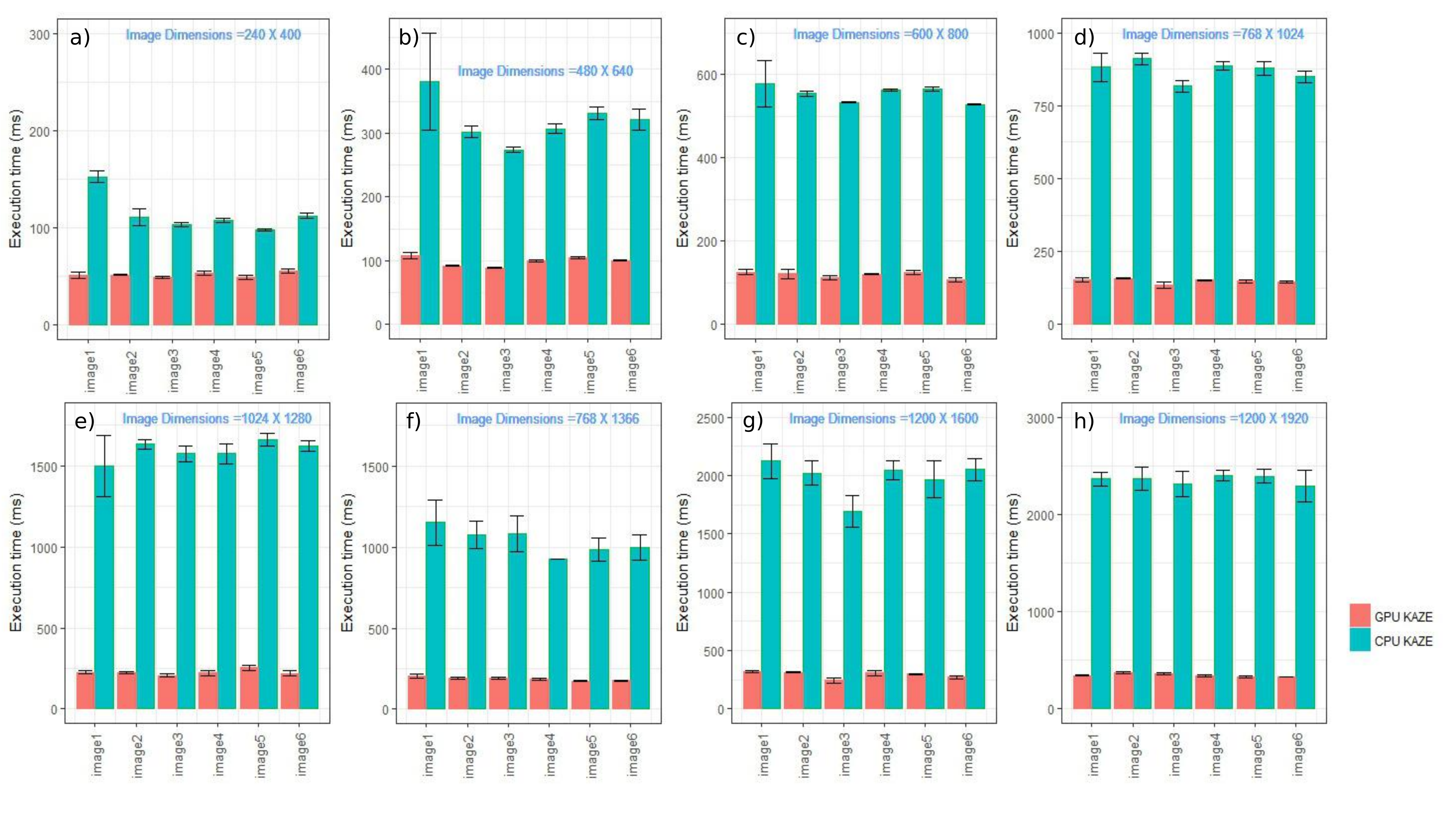}
\caption{Runtimes of  CPU-KAZE and GPU-KAZE for different scales}
\label{fig:Timing performance2}
\end{figure*}

\subsection{ Implementation of Feature Points Description}

\subsubsection{Implementation of Orientation Assignment}

To build the orientation, we divide the circle into segments each covering an angle
of $\pi/3$. We have defined two dimensional thread block with 121 threads ($11 \times 11$) such that each
thread computes the orientation of a keypoint. Information of all
neighborhood pixels around a keypoint in a circular region with radius of
$6\sigma$ ($\sigma$ is 1.5 times as the scale of the keypoint) is required to
compute orientation. The information of keypoint and their relative pixels that
have been computed is transferred to each thread using shared memory. We compute
Gaussian weighted first order derivatives centered at the keypoint for each sample
in the segments of a circular area. We represent the corresponding derivative
responses as points in vector space and is stored in GPU memory space. Finally,
the computed responses are summed over the sliding segments of the circle over an
angle $\pi/3$. The maximum response in the longest vector is the dominant
orientation.    

\subsubsection{Implementation of Feature Points Descriptor}

Computation of  MSURF-descriptors with the non-linear space is parallelized. We
use shared memory to accelerate the computation of descriptors and store the
derivatives responses over the squared regions. First order derivatives $L_{x}$
and $L_{y}$ are computed for detected feature at each scale $\sigma_{i}$ over
square region of $24\sigma_{i} \times 24 \sigma_{i}$ around keypoint. We divide
each region into $4 \times 4$ square sub-regions whose edge are $9\sigma_{i}$. Compute
the Gaussian weighted ($\sigma$ is 2.5 times as the scale of the keypoint)
derivative responses in each subregion around its center and summed into
descriptor vector. Compute Gaussian weighted ($\sigma$ is 1.5 times as the scale
of the keypoint) subregion vector around the keypoint over the mask of $4 \times 4$
region.  A 4-orientation derivative response is generated by calculating the
contribution of the orientation of each pixel to the orientation in a sub-region.
So we can obtain $4 \times 4 \times 4$ responses to form a 64-dimensional vector. We use one
thread to process a sub-region. In our allocation strategy, there are 16 threads
in a block to process 4 keypoints. A thread computes the weight of all the pixels in the sub-region and
transfer responses to shared memory. Then the processing results of the 16
threads could generate a 64-dimensional feature vector.

The number of keypoints decreases rapidly as the scale size $\sigma_{i}$ gradually increases
so does the number of threads which are used to process and compute information
of keypoints. A preprocessing before orientation assignment and keypoints descriptor is put
forward. In this preprocess, some information of images such as size, scale etc.,
and the memory address of feature vectors are calculated by CPU and stored in GPU global
memory. Because these data would not be accessed frequently and never modified,
they are stored in the constant memory which saves bandwidth and accelerates the
accessing speed. Moreover, all the images and information of keypoints
should be bound to several texture references. After pre-processing, the kernel
will be initialized.

To improve load balance, we adapted the following process.
\begin{enumerate}
  \item For a given keypoint, the location of the keypoint in image pyramid is calculated by each thread. 
\item To compute keypoint descriptor vectors and their corresponding orientations, we should reuse the information stored in constant memory.   
  \item the computed feature vectors should be stored in the appointed GPU memory addresses.
  \end{enumerate} 
In contrast to methods by which the scale space levels are processed one by one, all the image scales are computed at the same time to make full use of threads that are potentially starving. We reallocate the blocks and threads which process the keypoints
in different scale images concurrently. Consequently, this allocation
strategy ameliorates the load imbalance.

\section{Results and Discussion}
\label{sec:results}

\begin{figure}[htbp]
\centering
\includegraphics[width=0.7\textwidth]{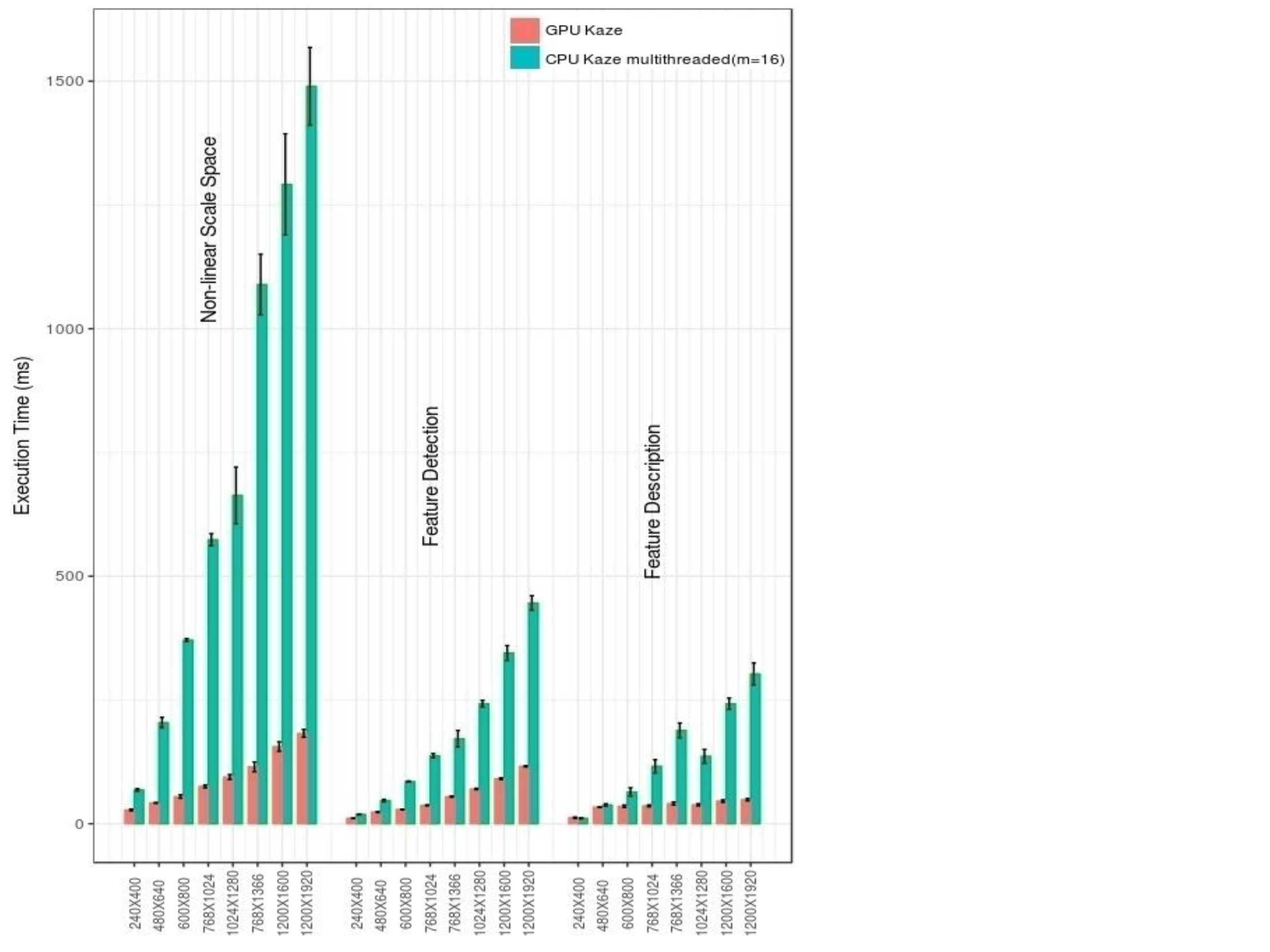}

\caption{Individual runtimes of key steps in CPU-KAZE and GPU-KAZE}
\label{fig:Timing performance3}
\end{figure}

\begin{figure*}[htbp]
\centering
\includegraphics[width=\textwidth]{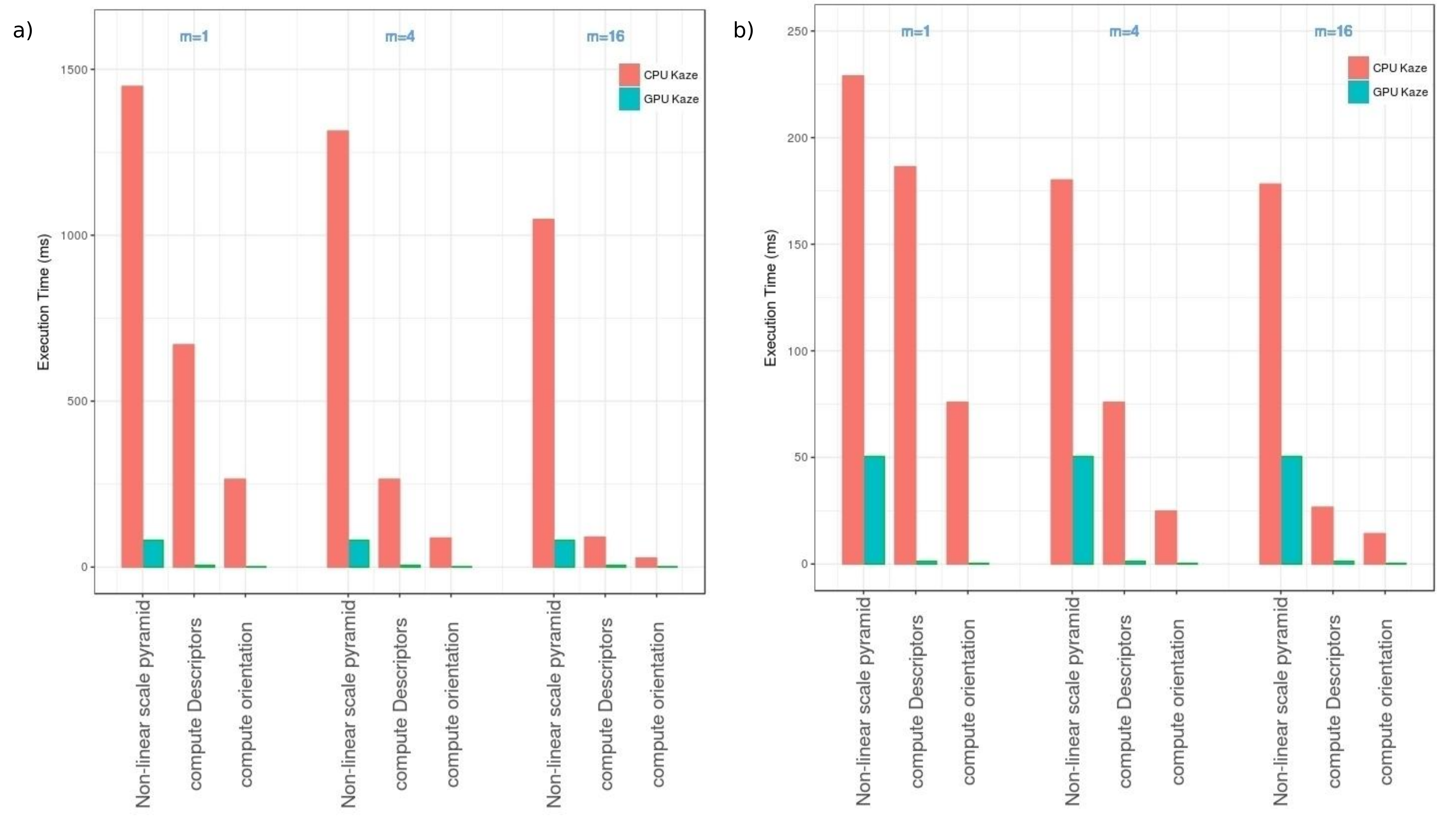}

\caption{Runtimes of multithreaded CPU-KAZE and GPU-KAZE for key tasks with respect to the image dimensions of  $1920 \times 1200 $ and  $480 \times 640$ respectively }
\label{fig:Timing performance4}
\end{figure*}

In this paper, we implement
the program on the GPU NVIDIA Geforce GTX TITAN X (MAXWELL). It has 12 GB of global memory
and 64 KB of constant memory. This GPU has 48KB shared memory and at most 65536
registers on each Streaming Multiprocessors (SM). Moreover, This GPU also has 32
blocks and 64 warps with 32 threads per warp on each SM. The CPU version of
KAZE is implemented on Intel(R) Xeon(R) CPU E5-2650 processor configured with 8
cores and 16 threads.

In~\autoref{fig:Timing performance}(a), we plot the runtimes for both the multithreaded CPU-KAZE with $m=16$ (where '$m$' denotes the number of threads) and our GPU-KAZE implementation for various image sizes. While the runtimes increase as a function of the image size, the rate of increase is significantly larger for the CPU version. The speedup of the GPU version is thus seen to improve with increasing image size. GPU-KAZE is around 10 times faster than the CPU-KAZE for the image of size  $1920 \times 1200$. In~\autoref{fig:Timing performance}(b), we have plotted the runtimes of CPU-KAZE with varying number of threads ($(m)=4,8,12,16$) with respect to increasing image dimensions. It is seen that the speedup does not scale well with increased thread count. The GPU version, however, will scale well with the number of CUDA cores in the GPU chip. 

For a given image size the number of keypoints will change depending on the complexity of the image and thus runtimes for images of the same size may change. In order to study the effect of image complexity, we created an image dataset of 8 different image dimensions ranging from $240 \times 400$ (lower image dimension) to $1920 \times 1200$ (Higher image dimension) with 6 images of varying complexity in each dimension. 
In~\autoref{fig:Timing performance2}, we have plotted the average time taken by CPU-KAZE and GPU-KAZE for each image in this dataset grouped by their dimension. Although there is now variance in the runtime for an image of a given size, the speedup factors observed earlier still seem to hold. 

In~\autoref{fig:Timing performance3}, we assess the runtimes with respect to the individual key steps involved in the KAZE algorithm. We can notice almost 10 times faster performance of GPU-KAZE than CPU-KAZE for each step as the image dimension increases. We can also observe that building non-linear scale space consumes most of the time when compared to feature detection (computes multiscale derivatives and keypoints) and feature description (computes descriptors). 

Within these three major steps, construction of the Nonlinear scale space pyramid, computation of the keypoints orientation and descriptor calculation are the substeps with the largest computational burden. 
The runtimes of GPU-KAZE and multithreaded CPU-KAZE (with variation in the number of threads $(m) = 1,4,16$)  for these subtasks are plotted in~\autoref{fig:Timing performance4}. Two different image dimensions were considered. It is clearly seen that better speedups are obtained especially for the nonlinear pyramid construction step for the image with larger dimensions. The speedup factor for the smaller and large dimension is listed in~\autoref{table:comp_SD} and~\autoref{table:comp_HD} respectively.

\begin{figure}[htbp]
\centering
\includegraphics[width=0.5\textwidth]{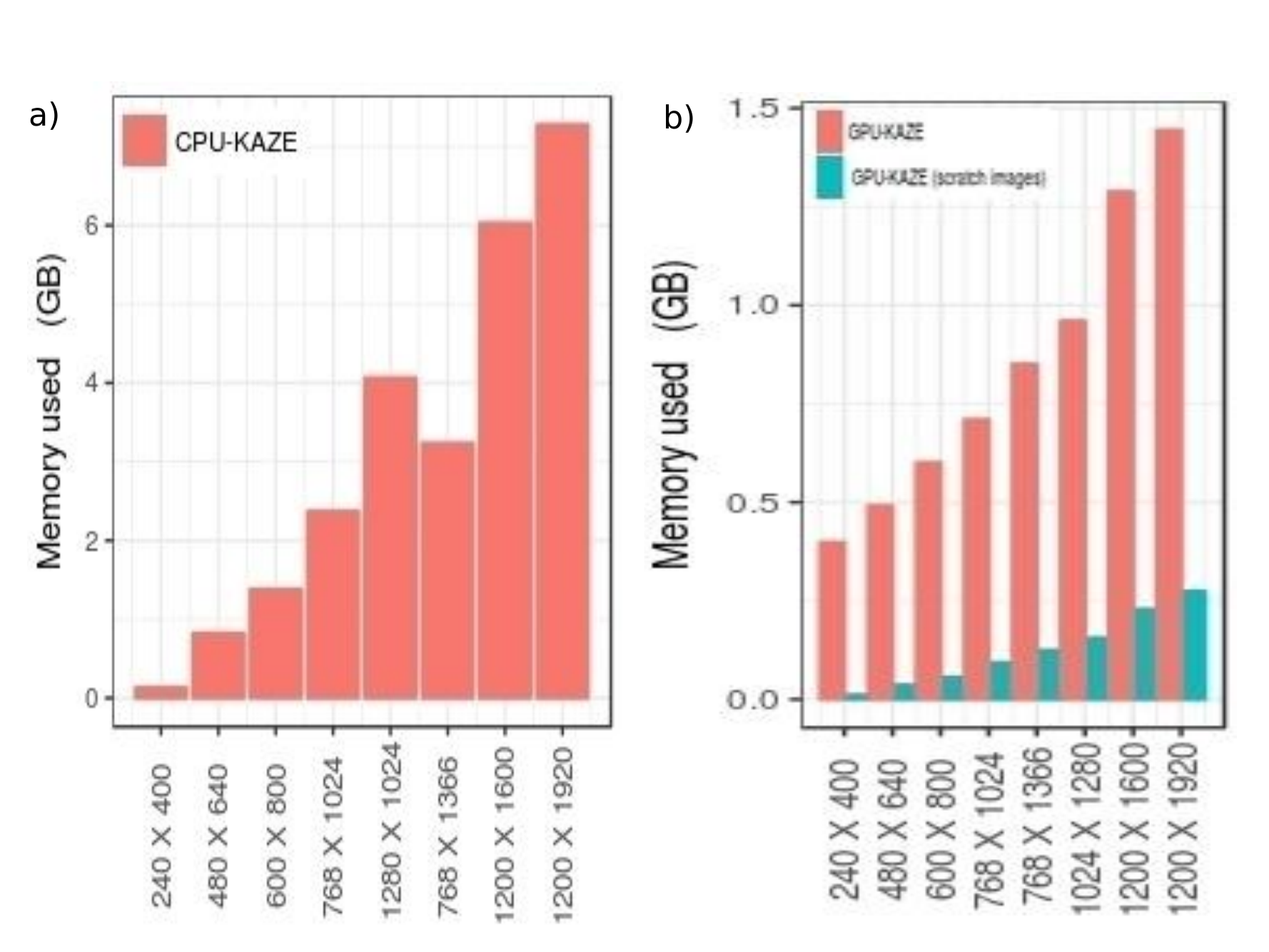}
\caption{Memory foot print of CPU-KAZE and GPU-KAZE for different scales}
\label{fig:Memory foot print}
\end{figure} 

\begin{figure*}[htbp]
\centering
\includegraphics[width=\textwidth]{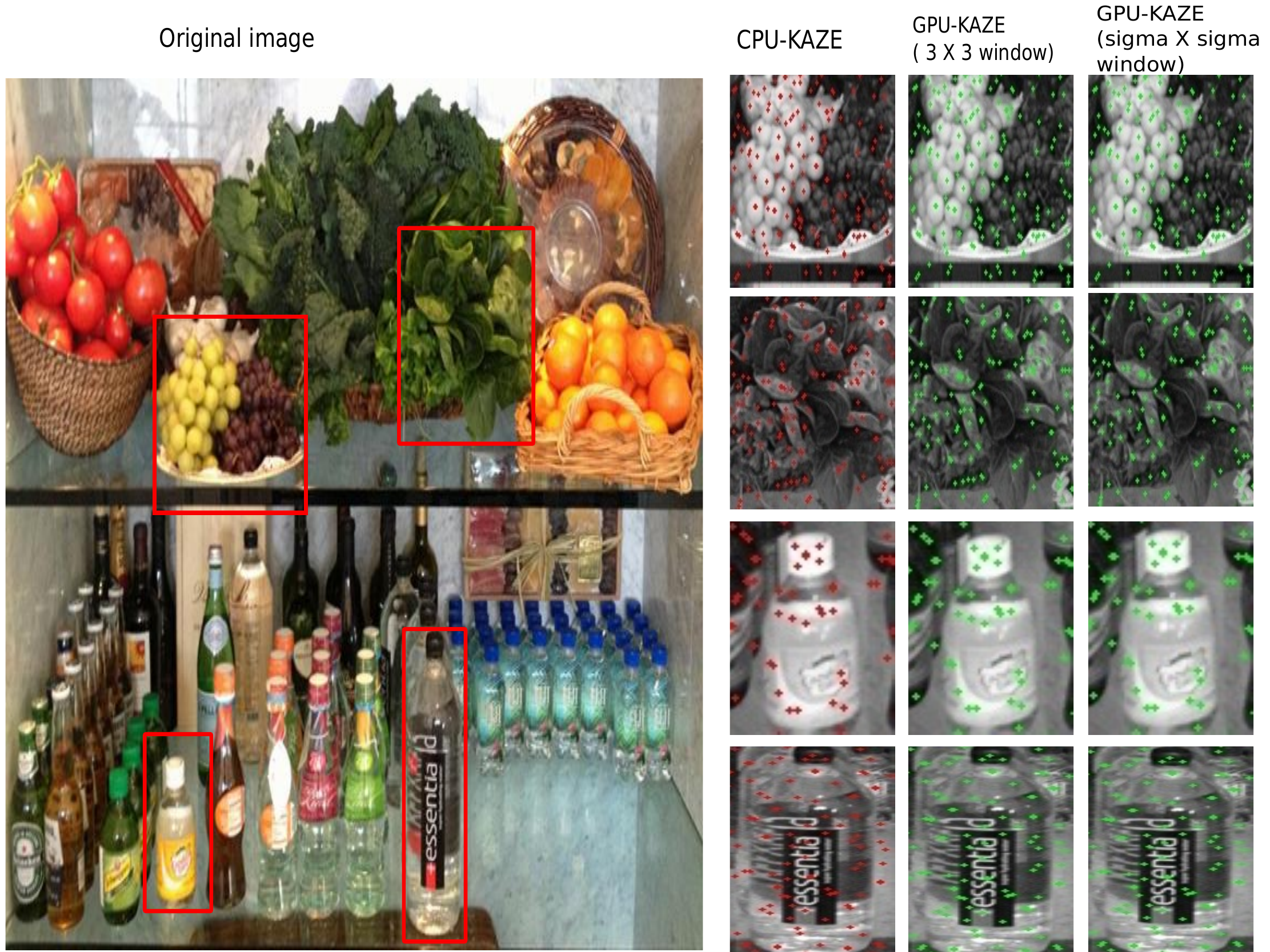}
\caption{Keypoints obtained from  CPU-KAZE and GPU-KAZE }
\label{fig:Key points}
\end{figure*}

Efficient memory optimization is also a key factor that affects the performance capabilities of any algorithm. The impact on performance by effective utilization, optimization, and reuse of heap space can be observed as image dimensions increases or when a huge number of keypoints are extracted. In~\autoref{fig:Memory foot print}, we plot the memory utilization of CPU-KAZE that has been analyzed using Valgrind tool whereas GPU-KAZE is analyzed using NVIDIA profiler. Memory utilized by CPU-KAZE is shown in~\autoref{fig:Memory foot print}(a). In~\autoref{fig:Memory foot print}(b), we also plot memory utilized by GPU-KAZE and memory utilized by scratch images say $L_{step}$ (represented by the green colored bar chart) with varying image dimensions. We can notice that the memory utilized by CPU-KAZE for the larger image dimension ($1200\times 1920$) is around twenty times higher than the memory utilized by CPU-KAZE for the lower image dimension ($480\times 640$) whereas memory utilized by GPU-KAZE for the larger image dimension ($1200\times 1920$) is around three times higher than the memory utilized by GPU-KAZE for the lower image dimension ($480\times 640$) implying better memory optimization in the case of GPU-KAZE when compared with CPU-KAZE.  
\begin{table}[htbp]
\caption{GPU-KAZE over multithreaded CPU- ($m=$ denotes number of CPU threads) speedup factor for the image size ~$480\times 640$ }
\label{table:comp_SD}
\centering
\begin{tabular}{l || c | c | c }
 & $m=1$ & $m=4$ & $m=16$ \\ \hline\hline
Construct nonlinear scale space & 5 & 3.5 & 3.5 \\ \hline
Compute feature descriptors & 130 & 60 & 20\\ \hline
Compute keypoint orientations & 200 & 75 & 20
\end{tabular}
\end{table}

\begin{table}[htbp]
\caption{GPU-KAZE over multithreaded CPU-KAZE ($m=$ denotes number of CPU threads) speedup factor for the image size ~$1200\times 1920$}
\label{table:comp_HD}
\centering
\begin{tabular}{l || c | c | c }
 & $m=1$ & $m=4$ & $m=16$ \\ \hline\hline
Construct nonlinear scale space & 18 & 16 & 13 \\ \hline
Compute feature descriptors & 120 & 45 & 18\\ \hline
Compute keypoint orientations & 200 & 65 & 20
\end{tabular}
\end{table}
In~\autoref{fig:Key points}, we compare the extracted keypoints of CPU-KAZE and GPU-KAZE implementations. The CPU-KAZE implements an approximate serial procedure by first checking for maxima on a $3 \times 3$ window instead of the $\sigma \times \sigma$ window. We have implemented both this approximate procedure and also the exact procedure. In the GPU version, no runtime penalty was observed by implementing the exact procedure instead of the approximate one. We have considered an image with dimension $480 \times 640$ to check the accuracy of the keypoint extraction for the GPU-KAZE as compared to the CPU-KAZE. We have obtained a similar number of keypoints (i.e., around 2200 keypoints are extracted) for both CPU-KAZE and GPU-KAZE. We have noticed relatively less number of keypoints when using the exact procedure in comparison to the number of keypoints obtained using CPU-KAZE approximate procedure.        

\section{Conclusion}
\label{sec:conclusion}
The computational cost for building non-linear scale space, feature extraction, and feature description of GPU-KAZE are in the ratio $4:3:1$ respectively. From the keen analysis of the results obtained from GPU-KAZE, we can notice that building nonlinear scale space and feature detection are the key tasks that are computationally intensive. To improve the nonlinear scale pyramid construction, new approaches to speed the solution of the Perona Malik PDE are needed. We are investigating schemes that have been utilized to perform nonlinear diffusion filtering~\cite{Chen2015,Hu2014,Chen2016} and can solve Partial Differential Equations (PDE)~\cite{Katzourakis2017} more efficiently. The feature detection computations rely on spatial derivative calculation in a three dimensional space. Utilizing the spatial locality features in GPU texture memory is one way to improve these.


\ifCLASSOPTIONcompsoc
  \section*{Acknowledgments}
\else
  \section*{Acknowledgment}
\fi

RB and RSH acknowledge funding support from Innit Inc. Consultancy grant CNS/INNIT/EE/P0210/1617/007 and High Performance Computing Lab support from Mr. Sudeep Banerjee.



%

\bibliographystyle{IEEEtran} 
\bibliography{hpc.bib}

\begin{thebibliography}{10}
\providecommand{\url}[1]{#1}
\csname url@samestyle\endcsname
\providecommand{\newblock}{\relax}
\providecommand{\bibinfo}[2]{#2}
\providecommand{\BIBentrySTDinterwordspacing}{\spaceskip=0pt\relax}
\providecommand{\BIBentryALTinterwordstretchfactor}{4}
\providecommand{\BIBentryALTinterwordspacing}{\spaceskip=\fontdimen2\font plus
\BIBentryALTinterwordstretchfactor\fontdimen3\font minus
  \fontdimen4\font\relax}
\providecommand{\BIBforeignlanguage}[2]{{%
\expandafter\ifx\csname l@#1\endcsname\relax
\typeout{** WARNING: IEEEtran.bst: No hyphenation pattern has been}%
\typeout{** loaded for the language `#1'. Using the pattern for}%
\typeout{** the default language instead.}%
\else
\language=\csname l@#1\endcsname
\fi
#2}}
\providecommand{\BIBdecl}{\relax}
\BIBdecl

\bibitem{Alcantarilla2012}
P.~Alcantarilla, A.~J.Davison, and A.~Bartoli, ``{Kaze Features},''
  \emph{Procedings of the British Machine Vision Conference}, vol. 7577 LNCS,
  no.~6, pp. 13.1--13.11, 2013.

\bibitem{Lindeberg1998}
T.~Lindeberg, ``{Feature Detection with Automatic Scale Selection},''
  \emph{International Journal of Computer Vision}, vol.~30, no.~2, pp. 79 --
  116, 1998.

\bibitem{Bianco2015}
S.~Bianco, D.~Mazzini, D.~P. Pau, and R.~Schettini, ``{Local detectors and
  compact descriptors for visual search: A quantitative comparison},''
  \emph{Digital Signal Processing: A Review Journal}, vol.~44, no.~1, pp.
  1--13, 2015.

\bibitem{Gauglitz2011}
S.~Gauglitz, T.~Hollerer, and M.~Turk, ``{Evaluation of interest point
  detectors and feature descriptors for visual tracking},'' \emph{International
  Journal of Computer Vision}, vol.~94, no.~3, pp. 335--360, 2011.

\bibitem{Sanna2014}
A.~Sanna and F.~Lamberti, ``{Advances in target detection and tracking in
  forward-looking infrared (FLIR) imagery},'' \emph{Sensors (Switzerland)},
  vol.~14, no.~11, pp. 20\,297--20\,303, 2014.

\bibitem{Fung2008}
J.~Fung and S.~Mann, ``{Using graphics devices in reverse: GPU-based Image
  Processing and Computer Vision},'' \emph{2008 IEEE International Conference
  on Multimedia and Expo, ICME 2008 - Proceedings}, pp. 9--12, 2008.

\bibitem{Seung2008}
I.~P. Seung, S.~P. Ponce, J.~Huang, Y.~Cao, and F.~Quek, ``{Low-cost,
  high-speed computer vision using NVIDIA's CUDA architecture},''
  \emph{Proceedings - Applied Imagery Pattern Recognition Workshop}, 2008.

\bibitem{Coates2013}
A.~Coates, B.~Huval, T.~Wang, D.~Wu, and A.~Y. Ng, ``{Deep learning with COTS
  HPC systems},'' \emph{Proceedings of The 30th International Conference on
  Machine Learning}, pp. 1337--1345, 2013.

\bibitem{Lowe2004}
D.~G. Lowe, ``{Distinctive image features from scale-invariant keypoints},''
  \emph{International Journal of Computer Vision}, vol.~60, no.~2, pp. 91--110,
  2004.

\bibitem{Mikolajczyk2005}
K.~Mikolajczyk, K.~Mikolajczyk, C.~Schmid, and C.~Schmid, ``{A performance
  evaluation of local descriptors},'' \emph{IEEE Transactions on Pattern
  Analysis and Machine Intelligence}, vol.~27, no.~10, pp. 1615--1630, 2005.

\bibitem{Rosten2006}
E.~Rosten and T.~Drummond, ``{Machine learning for high-speed corner
  detection},'' \emph{Lecture Notes in Computer Science (including subseries
  Lecture Notes in Artificial Intelligence and Lecture Notes in
  Bioinformatics)}, vol. 3951 LNCS, pp. 430--443, 2006.

\bibitem{Calonder2010}
M.~Calonder, V.~Lepetit, C.~Strecha, and P.~Fua, ``{BRIEF: Binary robust
  independent elementary features},'' \emph{Lecture Notes in Computer Science
  (including subseries Lecture Notes in Artificial Intelligence and Lecture
  Notes in Bioinformatics)}, vol. 6314 LNCS, no. PART 4, pp. 778--792, 2010.

\bibitem{Leutenegger2011}
S.~Leutenegger, M.~Chli, and R.~Y. Siegwart, ``{BRISK: Binary Robust invariant
  scalable keypoints},'' \emph{Proceedings of the IEEE International Conference
  on Computer Vision}, pp. 2548--2555, 2011.

\bibitem{Rublee2011}
E.~Rublee, V.~Rabaud, K.~Konolige, and G.~Bradski, ``{ORB: An efficient
  alternative to SIFT or SURF},'' \emph{Proceedings of the IEEE International
  Conference on Computer Vision}, pp. 2564--2571, 2011.

\bibitem{Lehiani2016}
Y.~Lehiani, M.~Preda, M.~Maidi, and F.~Ghorbel, ``{Object identification and
  tracking for steady registration in mobile augmented reality},'' \emph{IEEE
  2015 International Conference on Signal and Image Processing Applications,
  ICSIPA 2015 - Proceedings}, pp. 54--59, 2016.

\bibitem{Gao2017}
\BIBentryALTinterwordspacing
X.~Gao, W.~Li, M.~Loomes, and L.~Wang, ``{A fused deep learning architecture
  for viewpoint classification of echocardiography},'' \emph{Information
  Fusion}, vol.~36, pp. 103--113, 2017. [Online]. Available:
  \url{http://dx.doi.org/10.1016/j.inffus.2016.11.007}
\BIBentrySTDinterwordspacing

\bibitem{Camargo2014}
A.~Camargo, D.~Papadopoulou, Z.~Spyropoulou, K.~Vlachonasios, J.~H. Doonan, and
  A.~P. Gay, ``{Objective definition of rosette shape variation using a
  combined computer vision and data mining approach},'' \emph{PLoS ONE},
  vol.~9, no.~5, 2014.

\bibitem{Zhai2014}
Y.~Zhai, Y.~S. Ong, and I.~W. Tsang, ``{The emerging Big dimensionality},''
  \emph{IEEE Computational Intelligence Magazine}, vol.~9, no.~3, pp. 14--26,
  2014.

\bibitem{Alcantarilla2013}
P.~Alcantarilla, J.~Nuevo, and A.~Bartoli, ``{Fast Explicit Diffusion for
  Accelerated Features in Nonlinear Scale Spaces},'' \emph{Procedings of the
  British Machine Vision Conference 2013}, pp. 13.1--13.11, 2013.

\bibitem{Pieropan2016}
\BIBentryALTinterwordspacing
A.~Pieropan, M.~Bj{\"{o}}rkman, N.~Bergstr{\"{o}}m, and D.~Kragic, ``{Feature
  Descriptors for Tracking by Detection: a Benchmark},'' 2016. [Online].
  Available: \url{http://arxiv.org/abs/1607.06178}
\BIBentrySTDinterwordspacing

\bibitem{Bay2008}
H.~Bay, A.~Ess, T.~Tuytelaars, and L.~{Van Gool}, ``{Speeded-Up Robust Features
  (SURF)},'' \emph{Computer Vision and Image Understanding}, vol. 110, no.~3,
  pp. 346--359, 2008.

\bibitem{Harvey1996}
R.~W. Harvey, A.~Bosson, and J.~A. Bangham, ``{A Comparison of Linear and
  Non-Linear Scale-Space Filters in Noise},'' \emph{Signal Processing VIII},
  vol.~1, no.~6, pp. 1777--1781, 1996.

\bibitem{Perona1990}
``{Scale-Space and Edge Detection Using Anisotropic Diffusion},'' pp. 629--639,
  1990.

\bibitem{Feng2016}
L.~Feng, Z.~Wu, and X.~Long, ``{Fast Image Diffusion for Feature Detection and
  Description},'' \emph{International Journal of Computer Theory and
  Engineering}, vol.~8, no.~1, pp. 58--62, 2016.

\bibitem{Grewenig2010}
S.~Grewenig, J.~Weickert, and A.~Bruhn, ``{From box filtering to fast explicit
  diffusion},'' \emph{In: Goesele M., Roth S., Kuijper A., Schiele B.,
  Schindler K. (eds) Pattern Recognition. DAGM 2010. Lecture Notes in Computer
  Science}, vol. 6376 LNCS, pp. 533--542, 2010.

\bibitem{Li2015}
Y.~Li, S.~Wang, Q.~Tian, and X.~Ding, ``{A survey of recent advances in visual
  feature detection},'' \emph{Neurocomputing}, vol. 149, no.~PB, pp. 736--751,
  2015.

\bibitem{Weickert2002}
\BIBentryALTinterwordspacing
J.~Weickert and H.~Scharr, ``{A Scheme for Coherence-Enhancing Diffusion
  Filtering with Optimized Rotation Invariance},'' \emph{Journal of Visual
  Communication and Image Representation}, vol.~13, no. 1-2, pp. 103--118,
  2002. [Online]. Available:
  \url{http://linkinghub.elsevier.com/retrieve/pii/S104732030190495X}
\BIBentrySTDinterwordspacing

\bibitem{Dalal2005}
N.~Dalal and B.~Triggs, ``{Histograms of oriented gradients for human
  detection},'' \emph{Proceedings - 2005 IEEE Computer Society Conference on
  Computer Vision and Pattern Recognition, CVPR 2005}, vol.~I, pp. 886--893,
  2005.

\bibitem{Huang2010}
H.~Huang, L.~Lu, B.~Yan, and J.~Chen, ``{A new scale invariant feature detector
  and modified SURF descriptor},'' \emph{Proceedings - 2010 6th International
  Conference on Natural Computation, ICNC 2010}, vol.~7, pp. 3734--3738, 2010.

\bibitem{Chen2015}
Y.~Chen and T.~Pock, ``{Trainable Nonlinear Reaction Diffusion: A Flexible
  Framework for Fast and Effective Image Restoration},'' \emph{IEEE
  Transactions on Pattern Analysis and Machine Intelligence}, vol.~39, no.~6,
  pp. 1256--1272, 2015.

\bibitem{Hu2014}
W.~Hu, R.~Hu, N.~Xie, H.~Ling, and S.~Maybank, ``{Image Classification Using
  Multiscale Information Fusion Based on Saliency Driven Nonlinear Diffusion
  Filtering},'' \emph{Image Processing, IEEE Transactions on}, vol.~23, no.~4,
  pp. 1513--1526, 2014.

\bibitem{Chen2016}
B.~Chen, X.-H. Zhou, L.-W. Zhang, J.~Wang, W.-Q. Zhang, and C.~Zhang, ``{A new
  nonlinear diffusion equation model for noisy image segmentation},''
  \emph{Advances in Mathematical Physics}, vol. 2016, 2016.

\bibitem{Katzourakis2017}
\BIBentryALTinterwordspacing
N.~Katzourakis, ``{Generalised solutions for fully nonlinear PDE systems and
  existence–uniqueness theorems},'' \emph{Journal of Differential Equations},
  vol. 263, no.~1, pp. 641--686, 2017. [Online]. Available:
  \url{http://dx.doi.org/10.1016/j.jde.2017.02.048}
\BIBentrySTDinterwordspacing

\end{thebibliography}

\end{document}